# Extracting Angina Symptoms from Clinical Notes Using Pre-Trained Transformer Architectures

Aaron S. Eisman, ScB[1,2], Nishant R. Shah, MD, MPH[2,3,4], Carsten Eickhoff, PhD[1,2,5], George Zerveas, MSc[1,5], Elizabeth S. Chen, PhD[1,2,3], Wen-Chih Wu, MD, MPH[2,3,4], Indra Neil Sarkar, PhD, MLIS[1,2,3,6]

[1]Center for Biomedical Informatics, Brown University, Providence RI;
[2]The Warren Alpert Medical School, Brown University, Providence, RI;
[3]School of Public Health, Brown University, Providence, RI;
[4]Division of Cardiology, Providence VA Medical Center, Providence, RI;
[5]Department of Computer Science, Brown University, Providence, RI;
[6]Rhode Island Quality Institute, Providence, RI

**Abstract**

*Anginal symptoms can connote increased cardiac risk and a need for change in cardiovascular management. This study evaluated the potential to extract these symptoms from physician notes using the Bidirectional Encoder from Transformers language model fine-tuned on a domain-specific corpus. The history of present illness section of 459 expert annotated primary care physician notes from consecutive patients referred for cardiac testing without known atherosclerotic cardiovascular disease were included. Notes were annotated for positive and negative mentions of chest pain and shortness of breath characterization. The results demonstrate high sensitivity and specificity for the detection of chest pain or discomfort, substernal chest pain, shortness of breath, and dyspnea on exertion. Small sample size limited extracting factors related to provocation and palliation of chest pain. This study provides a promising starting point for the natural language processing of physician notes to characterize clinically actionable anginal symptoms.*

**Introduction**

Angina pectoris is a constellation of symptoms that portends inadequate oxygenation of cardiac muscle due to either a decrease in coronary blood supply, an increase in myocardial oxygen demand, or both.[1] These symptoms are classically described as substernal chest pain that worsens with exertion and improves with rest or administration of nitroglycerin.[2] Concomitant shortness of breath is also frequently reported. The presence of anginal symptoms has been demonstrated to increase the likelihood of underlying atherosclerotic cardiovascular disease substantially and is a key input variable along with traditional cardiovascular risk factors in estimating patients' pretest probability of coronary artery disease (CAD). This pretest probability, in turn, informs the appropriateness of providers' referral of patients for cardiac testing to identify obstructive CAD.[3] The CAD Consortium Score is a commonly used tool to estimate the pretest probability of CAD.[4] In this score, the presence of typical versus atypical anginal symptoms connotes a more than three-fold increase in pre-test probability of CAD.

While traditional cardiovascular risk factors (e.g., age, sex, race, hypertension, hyperlipidemia, diabetes, and smoking status) are generally available within the electronic health record (EHR) as structured data, angina symptoms are typically recorded as unstructured natural language free-text descriptions within physician notes.[5] A prior study found that chest pain history is recorded as structured International Classification of Disease (ICD) codes in the EHR only half of the time.[6] This prevents symptoms from being incorporated into automated clinical decision support systems, limits the evaluation of guideline adherence, and makes it difficult to do large scale research about the prognostic ability of symptoms outside of clinical trials[4] and natural language processing (NLP) challenges.[7]

Prior work to extract both anginal and other symptoms from clinician-written natural language has primarily utilized complex annotation procedures and specialized software requiring both clinical and linguistic expertise.[6,8–11] In addition, portability assessments outside of the environment in which these methods have been developed show performance reduction.[12] Challenges include the generalized detection of negation,[13] interpretation of time expression,[14] and retraining to account for institution-specific differences.[15] However, clinical note content documenting similar events exhibit important similarities that should enable the implementation of NLP tasks across institutions.[12] The combination of significant upfront work to develop NLP tools and the likelihood that they will require testing and retraining to be applied to new environments has made them impractical to implement.[16] Therefore, it is necessary to explore methods that minimize the need for manual feature labeling by both clinical and linguistic experts.

Pre-trained transformer encoder architectures are large language models that address several of these shortcomings and have been shown to produce state-of-the-art results on NLP tasks.[17] Trained to perform language modeling (e.g., next word prediction) on large corpora of text, they extract contextual representations of words by considering all words in the input sequence at once and selectively focusing on specific parts of their context, based on several aspects such as syntactic structure and semantic similarity. Transformer encoders, which typically comprise hundreds of millions of parameters or more, can thus extract rich, general linguistic information from text and then be fine-tuned to specific tasks, including sequence classification, requiring several orders of magnitude fewer data than would otherwise be needed to train a model from scratch. This study aimed to apply a pre-trained transformer model for the task of angina symptom detection in clinical notes that were written by primary care physicians referring patients for cardiac stress testing.[18] The specific publicly available pre-trained neural network was built on the Bidirectional Encoder Representation from Transformers (BERT) language model. It was first pre-trained for language modeling on a large-scale biomedical corpus and then on publicly available medical notes.[19,20] This study tested the hypothesis that such a language model can be fine-tuned on a small number of annotated physician notes for the purpose of extracting clinically relevant symptom information associated with angina pectoris.

**Methods**

*Patient Population*
Consecutive patients (n=459) without known CAD referred for cardiac testing by primary care physicians were included in the study cohort. This population was expected to be enriched for cardiovascular symptoms, including angina and anginal equivalents (e.g., dyspnea on exertion) as they are a common impetus for cardiac stress test referral. Baseline cardiac risk factors were collected including demographics (e.g., age and sex, and race), comorbidities (e.g., diabetes), blood pressure (e.g., systolic and diastolic), medications (e.g., statins, aspirin, and anti-hypertension), and serum cholesterol (e.g., total and high-density lipoprotein). Patients with a known history of prior percutaneous coronary intervention or coronary artery bypass graft surgery were excluded. This study was approved by the Providence VA Medical Center Institutional Review Board.

*Angina Definitions*
Typical angina is pain or discomfort that is: (1) substernal (i.e., center of the chest), (2) provoked by exertion or emotional stress, and (3) relieved by rest or nitroglycerin. The presence of two out of three of these symptoms constitutes atypical angina, and pain is considered nonanginal when it only meets a single criterion.[21] The category of nonanginal pain for the purpose of risk estimation has been broadened clinically to include other non-specific symptoms of potential cardiac origin including vague chest discomfort, shortness of breath, and dyspnea on exertion.[3]

*Manual Note Annotation*
Medical notes are routinely organized into sections that indicate expected content. Clinical symptoms can most frequently be found in the history of present illness (HPI), review of systems (ROS), or assessment/plan (A/P). This study focused on the HPI section, which represents the interval medical history as expressed by the patient and distilled by the note writer. For each patient in the study cohort, the primary care note most proximal to the cardiac stress test order date was identified using the VA's Computerized Patient Record System, and the relevant sections were extracted. HPI sections were selected as one single continuous portion of text from within the note, the beginning of which was usually indicated by a section header (e.g., "HPI" or "Subjective") and the end of which was indicated by the start of another note section (e.g., "Medications" or "Objective").

Study data were collected and managed using instruments implemented in the Research Electronic Data Capture (REDCap) tool.[22,23] HPIs were stored in free text fields and reviewed from within REDCap. Each HPI was annotated using symptom fields regarding the positive, negative, or absent mention of (1) chest pain or discomfort, (2) substernal chest pain, (3) chest pain provoked by exertion or emotional stress, (4) chest pain relieved by rest or nitroglycerin, (5) shortness of breath, and (6) dyspnea on exertion (**Table 1)**. The definition of 'substernal' was broadened from the original Diamond and Forrester description[21] to include imprecise anatomical descriptions that are colloquially meant as substernal or for all practical purposes are more likely than not a description of substernal pain or discomfort. This includes left-sided, left anterior, and descriptions of heaviness, tightness, heartburn, and precordial pain. Definitions of exertion included almost any association of the pain with a description of the activity. Two examples that were deemed not enough activity to warrant exertion were "with standing" and "while washing the dishes." The emotional stress criteria were considered met with the direct connection made between anxiety or an emotionally stressful event to the onset of chest pain at the same time. A patient who generally reported being anxious or under a lot of stress (e.g., family, job, or homelessness) but did not make the explicit link to chest pain symptoms was not annotated as having emotional stress-induced chest pain.

**Table 1.** Annotation Guidelines

| Symptom | Positive Examples | Negative Examples | Ambiguous Cases |
|---|---|---|---|
| **Chest Pain or Discomfort** | Chest: pain, pressure, tightness, ache, discomfort, heaviness, sitting on, burning, something "not quite right"<br><br>Described as: sharp, dull, severe, mild, stabbing<br><br>Acronyms: cp, sscp | any negation of a positive | "The patient has had two episodes of chest pain in the past month. Patient denies chest pain today." → positive |
| **Substernal CP** | CP location: substernal, sternal, central, center, middle, left, anterior, precordial, epigastric, across<br><br>CP quality that implies location: pressure, tightness, heaviness, sitting on, radiates to the left arm/shoulder<br><br>Acronyms: sscp, ssc, ss, l, l ant | right anterior, right, another body part only (shoulder, neck, arm, jaw, leg) | |
| **CP Provoked by Exertion or Emotional Stress** | Provocation: exertion, stress, walking, running, stairs, shoveling, mowing lawn, not specifically deemed "negative" | (w/out, w. out, without) "positive"<br>OR<br>(at/while) rest, laying down, night, sitting, standing up, washing dishes | "with rest and exertion" → positive |
| **CP Relieved by Rest or Nitroglycerin** | Palliation: rest, stopping, sitting, laying down, taking a few breaths | "positive" with time modifier that is excessively long (e.g. > 1hr) | "relieved by rest or continuing to walk" → positive |
| **Shortness of Breath** | Synonyms: shortness of breath, sob, dyspnea, any mention of breathing difficulty<br><br>Acronym: sob | any negation of a positive | |
| **Dyspnea on Exertion** | Defined with respect to dyspnea as exertional chest pain is defined with respect to chest pain.<br><br>Acronym: doe | | |

**CP – Chest Pain**

It was not uncommon for there to be internal inconsistencies within a single note for these categories. For example, the HPI may state that a patient experienced exertional chest pain symptoms recently but then go on to state that the patient had no chest pain at the time of the visit. Similarly, patients sometimes report that pain occurs both at rest and with exertion. Cases with both a positive and negative mention of a label were considered positive mentions unless the positive case was stated to have been resolved for a clearly explained reason. It took an average of two minutes to annotate each HPI for these symptoms. All annotations were performed by [ASE] and then reviewed by [NRS]. Disagreements were adjudicated between the two annotators with the help of [INS]. The final reference standard was the result of a consensus between the three coders.

*Transfer Learning with Clinical BERT*
BioBERT is a language representation model trained on both general (Wikipedia and BooksCorpus) and domain-specific (PubMed Abstracts and PMC Full-text articles) text.[24] The model has been fine-tuned on a language modeling task using discharge summaries from the Beth Israel Deaconess Medical Center Medical Information Mart for Intensive Care III database (MIMIC III). Bio+Discharge Summary BERT is publicly available within the Python HuggingFace transformers library as part of the BertForSequencClassification transformer class.[19,25] Bio+Discharge Summary BERT was chosen as a domain-appropriate language model for the downstream task of detecting anginal symptoms, which was cast as a classification task.

HPI text from each primary care note was embedded using Bio+Discharge Summary BERT. Each embedding was prepended with a [CLS] and padded with [PAD] tokens until reaching 512 total tokens, the maximum allowed by BERT models. Attention masks were constructed to enable batch processing of sequences by identifying which tokens should be attended to and which should be ignored as padding. Separate models were constructed to identify each of the six symptoms. Models were configured to identify three classes including positive, negative, and absent mentions of the symptom. The training was performed over 2, 4, 8, 16, 32, 64, and 128 epochs with a batch size of 8. The batch size was selected as the maximum number of 512 token embeddings that can be processed on one 12 GB GPU at a time. Data was split into training, validation, and testing sets (80/10/10) and then cross-validated ten-fold such that all data was represented in a collective validation and a testing set exactly once each. Matthews Correlation Coefficient (MCC) was calculated on pooled validation sets and plotted versus the number of training epochs for each symptom. The number of training epochs corresponding to the maximum of each of these curves was selected as the number of epochs to be used for reporting aggregate performance on the test set.

*Evaluation*
For each model, we reported the aggregate performance on the test set, after previously optimizing the number of training epochs based on aggregate performance on the validation set.

In order to assess the performance of detecting the positive presence of symptoms within the notes, the confusion matrices for each symptom were reduced to two-by-two such that the absent and negative classes were collapsed into a single class. Precision, recall, $F_1$-score, specificity, and Matthews Correlation Coefficient (MCC) were calculated. MCC was chosen in addition to the other information retrieval evaluation metrics that are not informed by true negative cases.[26] This was accomplished by balancing the ratios of the four quadrants of the binary confusion matrix (i.e., true positives, false positives, true negatives, and false negatives).

*Computational Resources*
This research was conducted using computational resources and services at the Center for Computation and Visualization at Brown University. 12GB NVIDIA GeForce Titan V and Tesla P100 GPUs were used for computation. Each epoch had a runtime of approximately 17-26 seconds (Titan vs P100 respectively) for a total of between 2.5 minutes to 18 minutes to fine-tune one model on the Titan V (8 and 64 epochs respectively).

**Results**

*Patient Characteristics*
After annotation of HPIs extracted from 459 consecutive patients without known CAD referred for cardiac testing by primary care physicians, 243 patients were noted to be experiencing chest pain and 205 were noted to be experiencing shortness of breath (SOB). Baseline clinical characteristics are reported by chest pain class (**Table 2**).

The median age of all patients was 66 years [57-71]. Patients were predominantly male (92%) and self-identified as white race (88%). Thirty percent had a history of diabetes. About half of the patients were on statin (58%), aspirin (44%), or antihypertensive (58%) therapies. Ten-year pooled cohort equation risk of developing atherosclerotic cardiovascular disease based on traditional risk factors was 18.1% (9.9-27.8).[5]

**Table 2.** Patient Characteristics

| Characteristic | All (N = 459) | Typical Chest Pain (N = 24) | Atypical Chest Pain (N = 73] | Non-Specific Symptoms (N = 255) | Asymptomatic (N = 107) |
|---|---|---|---|---|---|
| **Age (years)** | 66 [57-71] | 66.5 [57.5-71] | 62 [54-71] | 66 [59-70] | 65 [58-71] |
| **Sex (male)** | 423 (92%) | 24 (100%) | 65 (89%) | 231 (91%) | 103 (96%) |
| **Race (white)** | 406 (88%) | 22 (92%) | 64 (88%) | 230 (90%) | 90 (84%) |
| **Diabetes** | 137 (30%) | 9 (38%) | 16 (22%) | 78 (31%) | 34 (32%) |
| **Smoker (current)** | 94 (20%) | 3 (13%) | 15 (21%) | 51 (20%) | 25 (23%) |
| **SBP (mmHg)** | 132 [124-141] | 130 [121.8-141] | 129 [121-140] | 132 [124-141] | 132 [124-142] |
| **DBP (mmHg)** | 78 [73-82] | 76 [74-82.3] | 76 [70-83] | 78 [73-82] | 78 [74-82] |
| **Cholesterol** | | | | | |
| Total (mg/dL) | 179 [153-213] | 191 [159-216.5] | 179 [153-227] | 181 [153-212] | 177 [147.5-206.5] |
| HDL (mg/DL) | 44 [37-54] | 44 [39.5-47.5] | 46 [37-53] | 45 [37-54] | 43 [38-50.5] |
| **Chest Pain** | 243 (53%) | 24 (100%) | 73 (100%) | 146 (57%) | – |
| Substernal | 179 (39%) | 24 (100%) | 65 (89%) | 90 (35%) | – |
| Exertional | 108 (24%) | 24 (100%) | 65 (89%) | 7 (19%) | – |
| Improves with Rest | 42 (9%) | 24 (100%) | 16 (22%) | 2 (1%) | – |
| **Shortness of Breath** | 205 (45%) | 16 (67%) | 35 (48%) | 154 (60%) | – |
| Exertional | 155 (34%) | 16 (67%) | 29 (40%) | 110 (43%) | – |
| **Pharmacotherapy** | | | | | |
| **Statin Therapy** | 265 (58%) | 12 (50%) | 43 (59%) | 149 (58%) | 61 (57%) |
| High intensity | 121 (26%) | 7 (29%) | 21 (29%) | 67 (26%) | 26 (24%) |
| Moderate intensity | 134 (29%) | 5 (21%) | 20 (27%) | 77 (30%) | 32 (30%) |
| Low intensity | 10 (2%) | – | 2 (3%) | 5 (2%) | 3 (3%) |
| **Aspirin** | 201 (44%) | 11 (46%) | 28 (38%) | 112 (44%) | 50 (47%) |
| **Anti-HTN** | 264 (58%) | 13 (54%) | 33 (45%) | 146 (57%) | 72 (67%) |
| **Cardiovascular Risk** | | | | | |
| **10-Year ASCVD Risk**[5] | 18.1% [9.9-27.8] | 20.9% [9.1-27.9] | 14.3 [7.5-22.8] | 18.4 [10.7-28.0] | 20.5% [9.9-28.6] |
| **CAD Consortium Pretest Probability**[4] | | | | | |
| High | – | – | – | – | – |
| Moderate | 389 (85%) | 22 (92%) | 58 (79%) | 214 (84%) | 95 (89%) |
| Low | 70 (15%) | 2 (8%) | 15 (21%) | 41 (16%) | 12 (11%) |

**SBP** – systolic blood pressure, **DBP** – diastolic blood pressure, **HDL** – high-density lipoprotein, **HTN** – hypertension, **ASCVD** – atherosclerotic cardiovascular disease, **CAD** - coronary artery disease

*Note Annotations*
Notes were annotated for chest pain and shortness of breath symptoms. The majority of HPIs contained information about chest pain (77%) and shortness of breath (64%). Of the notes that contained positive mentions of chest pain, three quarters mentioned pain location while a minority described pain provocation (34%) or palliation (12%). Positive or negative mentions of shortness of breath were in 64% of HPIs, more than half of which specifically mentioned the presence or absence of exertional symptoms. Overall, there were three times as many positive as negative symptom mentions.

Despite the classification of distinct presence and absence of symptoms, clinical documentation of anginal symptoms often described complex narratives that required interpretation in order to be correctly classified. One example of this was a patient whose chest heaviness and shortness of breath had been linked to the non-cardiac source of mold exposure and since moving apartments, the patient reports the chest heaviness as being fully resolved and shortness of breath improving. In this case, the note would be correctly classified as negative for substernal chest pain and positive for shortness of breath. However, a prior note (not part of this study) that hypothesized mold as a potential cause of the patient's symptoms would be considered positive for both. The final model classified the patient as being positive for both symptoms (correct for shortness of breath and incorrect for substernal chest pain).

*Anginal Symptom Extraction*
Embedded HPI sections were truncated to BERT's 512 token limit. The majority of tokenized notes fit within the limit resulting in only 3.3% (n = 16) being truncated. A Chi-Square test demonstrated that the frequency of positive symptoms within the truncated samples was in line with the overall dataset ($p = 0.95$).

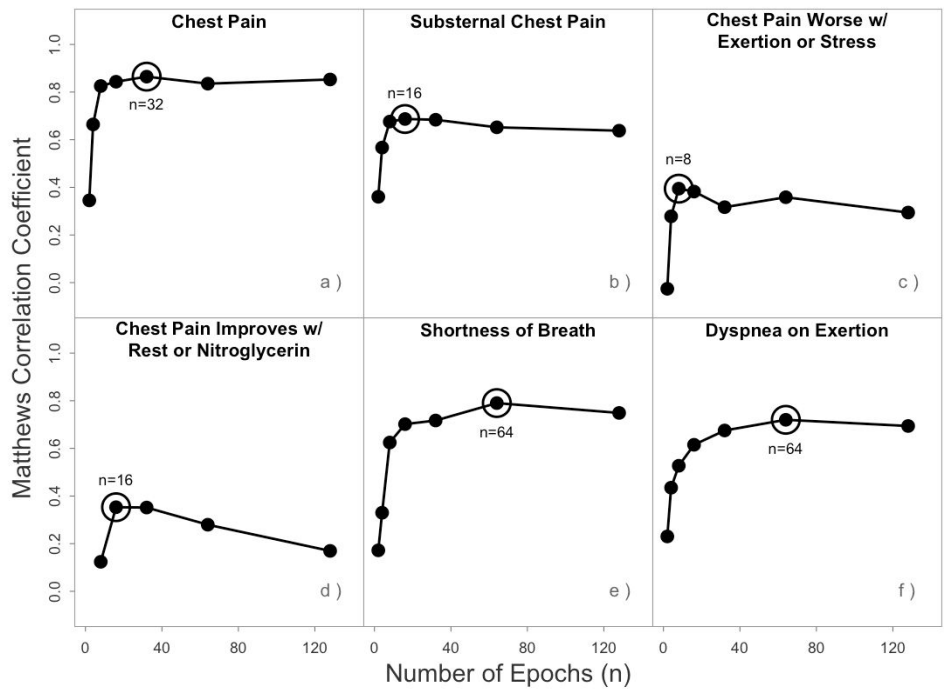

**Figure 1.** Validation Set Epoch Number Selection
Individual models for each symptom: (a) chest pain or discomfort, (b) substernal chest pain, (c) chest pain provoked by exertion or emotional stress, (d) chest pain relieved by rest or nitroglycerin, (e) shortness of breath, and (f) dyspnea on exertion, were trained using a range of epochs. Model performance (Matthews Correlation Coefficient) was evaluated as a function of epochs. The best performing number of epochs for each symptom was selected as the final model parameter for evaluation on held-out testing data.

BioBERT+Discharge Summary models were each fine-tuned to detect one of six different anginal symptoms using the labeled embedded HPI sections. The epoch number parameter was determined using an aggregated validation set. The graphs in **Figure 1** were inspected visually to determine the best early stopping parameter for each symptom. Parameter selection was straightforward given the limited number of epochs tested and the absence of local maximums observed. The chosen epoch number parameter was then used to test combined held-out test data from each of the ten cross-folds.

The final raw model performance is presented in a three-by-three confusion matrix for each symptom (**Table 3**). These include predicted annotation of absent, positive, and negative symptom mentions versus consensus manual annotation. For chest pain, the sequence classification model most common errors included negative identification of chest pain when it was absent (n = 14) and positive identification of chest pain when it was negative (n = 10). Negative assertions of chest pain characterization variables were under classified by the models. False positives of dyspnea on exertion were the most common breathing-related misclassification (n = 38).

**Table 3.** Confusion Matrix of Symptoms within HPI

| | | Adjudicated Manual Annotation | | | | | | | | |
|---|---|---|---|---|---|---|---|---|---|---|
| | | **Absent** | **+** | **–** | **Absent** | **+** | **–** | **Absent** | **+** | **–** |
| Predicted Annotation | | Chest Pain | | | SS CP | | | EX CP | | |
| | Absent | 90 | 4 | 1 | 223 | 26 | 4 | 274 | 47 | 23 |
| | + | 8 | 226 | 10 | 39 | 152 | 5 | 30 | 59 | 22 |
| | – | 14 | 9 | 97 | 0 | 0 | 0 | 0 | 2 | 2 |
| | | RE CP | | | SOB | | | DOE | | |
| | Absent | 389 | 29 | 12 | 136 | 7 | 2 | 245 | 14 | 6 |
| | + | 8 | 12 | 2 | 18 | 186 | 11 | 38 | 137 | 6 |
| | – | 5 | 1 | 1 | 12 | 12 | 75 | 7 | 3 | 3 |

SS CP – substernal chest pain, EX CP – exercise/stress-induced chest pain,
RE CP – chest pain improves with rest or nitroglycerin, SOB – shortness of breath, DOE – dyspnea on exertion

The confusion matrices were reduced to two-by-two tables where "Absent" and "–" assertions were collapsed into a single category to evaluate the model's performance to detect the presence or absence of symptom documentation. Precision, recall, F1, specificity, and Matthews Correlation Coefficient were calculated for each symptom and are reported in **Table 4**. Chest pain was extracted better than any other symptom (F1 = 0.936, MCC = 0.865). Chest pain subtype extraction performance ranged from good (substernal) to poor (improved with rest). Shortness of breath and dyspnea on exertion were extracted with good performance (F1 = 0.829, MCC = 0.695 and F1 = 0.818, MCC = 0.720 respectively).

**Table 4.** Model Evaluation Statistics

| | Chest Pain | SS CP | EX CP | RE CP | SOB | DOE |
|---|---|---|---|---|---|---|
| Precision | 0.926 | 0.776 | 0.532 | 0.545 | 0.865 | 0.757 |
| Recall/Sensitivity | 0.946 | 0.854 | 0.546 | 0.286 | 0.907 | 0.890 |
| $F_1$ | 0.936 | 0.812 | 0.539 | 0.375 | 0.886 | 0.818 |
| Specificity | 0.918 | 0.843 | 0.852 | 0.976 | 0.886 | 0.856 |
| MCC | 0.865 | 0.690 | 0.394 | 0.353 | 0.790 | 0.720 |
| **Training Epochs** | **32** | **16** | **8** | **16** | **64** | **64** |

SS CP – substernal chest pain, EX CP - exercise/stress-induced chest pain,
RE CP – chest pain improves with rest/nitroglycerin, SOB – shortness of breath, DOE – dyspnea on exertion
MCC – Matthews Correlation Coefficient

**Discussion**

This study examined the potential to use a pre-trained transformer architecture to extract anginal symptom information from clinical notes.[11] A publicly available BioBERT+Discharge language model was fine-tuned using HPI sections from 459 consecutive patients referred for cardiac testing. The models detected mention of chest pain with greater than 90 percent sensitivity and specificity. Chest pain location, shortness of breath, and dyspnea on exertion were similarly extracted with high sensitivity and specificity. These represent excellent starting points for the characterization of anginal symptoms from clinical notes of patients referred for cardiac testing.

Prior literature on the extraction of symptom information from clinical notes is sparse. A recent systematic review identified 27 articles related to use of NLP for clinical symptom detection and noted that study reproducibility is overall poor due to: (1) lack of detail about the patient population, (2) failure to report detailed performance, and (3) poorly described NLP methodology.[11] One study extracted chest pain and dyspnea symptoms with reported sensitivity on a small subset of their data, similar to the performance reported here.[9] However, no additional metrics, including false-positive rates, were provided for comparison, and methods lack details that would enable the system to be reproduced or implemented.

The performance of NLP systems to extract the provocation and palliation of chest pain has not been previously reported. The presented results indicate that the granular characterization of chest pain symptoms can be ruled out with high specificity. Reliable detection of these features requires additional consideration. The strength of the note for documentation is that it allows the clinician freedom to express complexity, ambiguity, and uncertainty of how symptoms are experienced and remembered. This leads to significant variability in how patient symptoms are ultimately documented. For example, it is useful to document that a patient has both experienced chest pain in the recent past and is pain-free at the time of the clinical encounter. However, these seemingly contradictory statements may be a challenge for a general language model attempting to determine if a patient has experienced chest pain at all. In addition, through the annotation process, it became clear that there is a wide range of ways that patients report, and clinicians document provocation and palliation of chest pain.

Fractional withholding of training data for both chest pain and substernal models indicate that performance is strongly related to the number of positive examples. Training models for these two symptoms with a comparable number of positive examples as the results reported for provocation and palliation of symptoms indicate similar performance. A small sample size for the provocation and palliation descriptions combined with narrative complexity is likely responsible for low sensitivity. Based on these findings, performance appears to plateau between 200 and 250 positive examples which may represent the necessary number to encompass the semantic variability of how chest pain is documented within the studied population.

The six classification objectives (i.e., angina symptoms) are not independent. Therefore, instead of training separate models for predicting each independently, it is possible to train a model that jointly predicts all objectives. Not only would deploying and maintaining a single model instead of six be more practical but this approach, known as multi-task learning, also offers theoretical benefits, such as the regularization of model weights.[27,28] The multi-task model shares the same architecture as the task-specific models, but with six distinct output layers instead of only one. The loss to be minimized during training is a weighted average of the six component losses. The multi-task model showed inferior predictive performance compared to the task-specific models. The potential reasons for these findings include: (a) the difficulty in stratifying the small training and test sets across 10 cross-validation folds in a way that preserves class balances and (b) the well-documented sensitivity of training to the weights of the individual tasks.[29]

Identifying clinically actionable anginal symptoms was the focus of this study. The developed models were designed to answer questions about patients that aid in the diagnosis[21] and determine the pretest probability[4] of coronary artery disease. Automating the extraction of anginal symptom information as presented in this study is potentially deployable as part of clinical decision support systems. It would enable health system-wide identification of high-risk patients and coordinate diagnostic cardiac testing for symptomatic ones.[3] In addition, automated symptom extraction is required for research on the relationship between anginal symptoms and downstream clinical testing (e.g., catheterization with and without percutaneous coronary intervention, coronary artery bypass graft surgery) and outcomes (e.g., myocardial infarction, cardiac death, and all-cause mortality).

Several limitations deserve mention that would need to be addressed for this work to be applied in a production environment. The process of identifying appropriate clinical notes for symptom extraction was performed manually and relies on institution-dependent EHR knowledge, including note types and physician relationship to the patient (e.g., primary care). In addition, relevant note section identification was also performed manually. This has been previously automated using institution-specific template headers,[30] and the authors are currently working on a generalizable solution to this problem. Finally, the transferability of these models to other clinical contexts and healthcare systems needs to be evaluated.

**Conclusion**

This study presents a promising method for detecting anginal and anginal-equivalent symptoms from clinical texts using pre-trained transformer architectures. The findings demonstrate that a generalized language model can be fine-tuned on a limited set of annotated physician notes to enable the extraction of clinically actionable anginal symptoms. The extracted symptoms align with data inputs for validated risk models and clinical guidelines and may allow the development of automated decision support and quality assessment. Additional work is required to assess the additional fine-tuning needed to apply these models in other clinical settings.

**Acknowledgments**

The research reported in this publication was supported in part by the National Institutes of Health grants F30LM013320, U54GM115677, and R25MH116440, and in part by Agency for Healthcare Research and Quality grant 5K12HS022998. The contents are solely the responsibility of the authors and do not necessarily represent the official views of the National Institutes of Health, Agency for Healthcare Research and Quality, the Department of Veterans Affairs, or the US government.